\pdfoutput=1
\documentclass[11pt]{article}
\usepackage{amssymb} % 用于显示对号 \checkmark
\usepackage{multirow}
\usepackage{graphicx}
\usepackage{array}
\usepackage{enumitem}
\usepackage{booktabs}  % 优化表格边框
\usepackage{makecell}  % 允许单元格内文本换行
\usepackage{pifont}
\usepackage[table]{xcolor}
\usepackage{colortbl}
\usepackage{hyperref}
\usepackage{amsmath}
\usepackage{arydshln}
\usepackage{tabularx}
\usepackage{xcolor} % 支持 \textcolor
\usepackage[most]{tcolorbox}
\usepackage{lmodern}
\usepackage{setspace}
\usepackage{longtable} % 请确保导言区加了这个包
\usepackage{color}

\usepackage[final]{acl}

\usepackage{times}
\usepackage{latexsym}

\usepackage[T1]{fontenc}
\usepackage[utf8]{inputenc}

\usepackage{microtype}

\usepackage{inconsolata}

\usepackage{graphicx}

\title{EmoStage: A Framework for Accurate Empathetic Response Generation via Perspective-Taking and Phase Recognition}

\author{
 \textbf{Zhiyang Qi\textsuperscript{1}},
 \textbf{Keiko Takamizo\textsuperscript{2,3}},
 \textbf{Mariko Ukiyo\textsuperscript{2,3}},
 \textbf{Michimasa Inaba\textsuperscript{1}}
\\
 \textsuperscript{1}The University of Electro-Communications,\\
 \textsuperscript{2}iDEAR Human Support Service,\\
 \textsuperscript{3}Japanese Organization of Mental Health and Educational Agencies
\\
 \texttt{\{qizhiyang,m-inaba\}@uec.ac.jp}
}

\begin{document}
\maketitle
\begin{abstract}

% 日益增长的心理健康护理需求引发了人们对人工智能驱动的咨询系统的兴趣。虽然大型语言模型 (LLM) 前景光明，但现有方法仍面临诸多挑战，例如对来访者心理状态和对话阶段的理解不足，对高质量训练数据的依赖，以及使用商业模型部署存在隐私风险。因此，我们提出了共情回复生成框架EmoStage，通过开源LLM自身的推理能力，旨在在无需训练数据的情况下，提高LLM的咨询回复生成能力。EmoStage通过引入换位思考，使模型能够从来访者的视角推测其情绪感受与支持需求，从而生成更贴近人心的回应。并结合阶段识别，判断当前咨询进程所处阶段，避免在不合时宜的时机提供建议，提升回复内容与对话流程的一致性。这种设计能够实现情感共鸣和情境感知的响应，同时适用于数据匮乏的语言环境，并降低隐私风险。我们在咨询数据稀缺的日语环境和合成数据丰富的中文环境中进行的实验表明，EmoStage提高了基础模型的响应质量，并取得了与数据驱动方法相比颇具竞争力的性能。

The rising demand for mental health care has fueled interest in AI-driven counseling systems. While large language models (LLMs) offer significant potential, current approaches face challenges, including limited understanding of clients' psychological states and counseling stages, reliance on high-quality training data, and privacy concerns associated with commercial deployment. To address these issues, we propose \textbf{EmoStage}, a framework that enhances empathetic response generation by leveraging the inference capabilities of open-source LLMs without additional training data. Our framework introduces perspective-taking to infer clients' psychological states and support needs, enabling the generation of emotionally resonant responses. In addition, phase recognition is incorporated to ensure alignment with the counseling process and to prevent contextually inappropriate or inopportune responses. Experiments conducted in both Japanese and Chinese counseling settings demonstrate that EmoStage improves the quality of responses generated by base models and performs competitively with data-driven methods.

\end{abstract}

\section{Introduction}

\begin{figure}[t!]
  \centering
  \includegraphics[width=\linewidth]{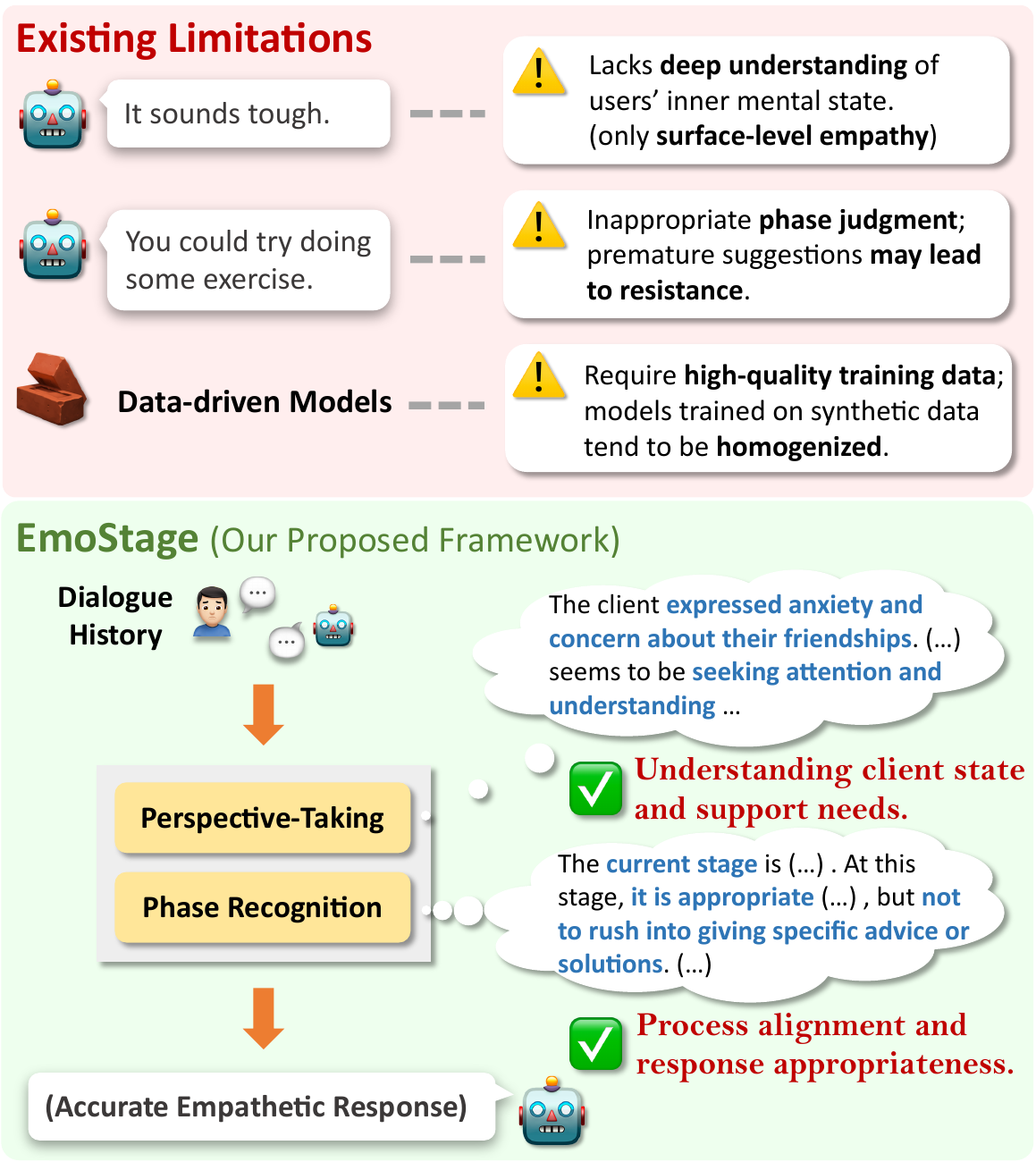}
  \caption{Based on existing limitations, we propose \textbf{EmoStage}, which aims to generate more accurate empathetic responses through \textbf{perspective-taking} and \textbf{phase recognition}.}
  \label{fig:fig_intro}
\end{figure}

% 全球每八人中就有一人患有精神障碍，抑郁症与焦虑症等疾病的发病率持续上升。然而，面对日益增长的心理健康服务需求，专业咨询资源依然严重不足，尤其在发展中国家和资源匮乏地区更为突出。近年来，在线心理咨询服务作为传统面对面咨询的补充，逐渐受到关注，能够缓解资源分布不均、费用高昂和地理限制等问题。与此同时，大型语言模型（LLM）的发展为心理咨询提供了新的解决思路。相较人类咨询师，LLM具备低成本、高可用性和易于部署的优势，特别适用于服务可及性受限的场景。已有研究开始探索构建“AI咨询师”，辅助或部分替代人类咨询师进行初步倾听、情绪支持与问题引导，为提升心理服务的可达性与效率提供了新的可能。

One in eight people worldwide suffers from a mental disorder \cite{world2022mental}, and the prevalence of conditions such as depression and anxiety continues to rise \cite{raising}. Despite growing demand for mental health services, access to professional counseling remains severely limited—particularly in developing countries and resource-constrained regions \cite{world2022mental}. 
In response, online psychological counseling has emerged as a promising supplement to traditional face-to-face therapy, helping to address challenges such as uneven resource distribution, high costs, and geographic barriers \cite{online}.

Meanwhile, advances in large language models (LLMs) have opened new possibilities for providing mental health support. Compared to human counselors, LLMs offer advantages such as lower cost, greater availability, and ease of deployment, making them particularly suitable for areas with limited access to care.
Recent studies have begun exploring the development of AI-based counselors \cite{ChatCounselor, chen-etal-2023-soulchat, qiu-etal-2024-smile}, which can assist or partially replace human therapists in tasks such as initial listening, emotional support, and problem exploration. These systems offer new opportunities to expand the accessibility and efficiency of psychological services \cite{support_1, support_2}.

% One in eight people worldwide suffers from a mental disorder \cite{world2022mental}, and the prevalence of conditions such as depression and anxiety continues to rise \cite{raising}. However, the availability of professional psychological counseling remains severely limited in the face of growing demand for mental health services—particularly in developing countries and resource-constrained regions, where this imbalance is even more pronounced \cite{world2022mental}. In recent years, online counseling services have gradually emerged, offering promising solutions to challenges faced by traditional face-to-face counseling, including limited availability, high costs, and geographic barriers \cite{online}. At the same time, the rapid advancement of large language models (LLMs) has opened new possibilities for mental health support. Recent studies have begun to explore the development of AI-based counselors powered by LLMs \cite{ChatCounselor, chen-etal-2023-soulchat, qiu-etal-2024-smile}, which can assist—or partially replace—human counselors in tasks such as initial listening, emotional support, and problem exploration, thereby addressing the expanding global need for psychological care \cite{support_1, support_2}.

% 尽管取得了一定进展，现有心理咨询对话系统仍存在诸多局限，如图1所示。首先，由于缺乏面对面交流中的非语言线索，单纯依赖对话文本建模的系统往往难以深入理解来访者的内在情绪与心理需求，生成的回应常流于表层共情。其次，心理咨询是一个多阶段、目标递进的过程，若系统无法识别当前所处阶段，可能在不恰当的时机提出建议，不仅无法有效支持来访者，反而可能引发情绪负担，削弱信任，甚至导致咨询中断。此外，心理咨询数据因隐私敏感性使得高质量的真实对话数据难以获取，因此很多研究采用合成数据训练语言模型。尽管在一定程度上缓解了数据不足问题，此类数据通常缺乏真实对话的复杂性与多样性，易造成模型输出趋于同质化。另一方面，尽管商业模型（如ChatGPT）具备强大的语言生成能力，但其闭源机制与数据处理的不透明性使得在涉及用户隐私的高敏感场景中应用受限。

Despite recent progress, existing mental health counseling dialogue systems still exhibit several limitations, as illustrated in Figure~\ref{fig:fig_intro}. First, due to the absence of non-verbal cues present in face-to-face interactions, systems that rely solely on dialogue-only textual modeling often struggle to fully understand the client’s underlying emotions and psychological needs, which may result in responses that remain at the level of superficial empathy. 
Second, many systems lack awareness of the counseling process's multi-stage structure. Psychological counseling is a goal-oriented process that unfolds in distinct stages \cite{phase}. Without recognizing the current stage, systems may provide advice at inappropriate times, potentially increasing emotional burden, undermining trust, or even causing the client to disengage from the session \cite{phase_wrong_1, phase_wrong_2}.
Moreover, the privacy-sensitive nature of counseling makes it difficult to collect high-quality real-world dialogue data. As a result, many studies rely on synthetic data to train language models \cite{chen-etal-2023-soulchat, zheng-etal-2023-augesc, qiu-etal-2024-smile, lee-etal-2024-cactus}. While synthetic data helps address data scarcity, it often lacks the complexity and diversity of authentic interactions \cite{zheng-etal-2024-self}, leading to homogeneous and less nuanced model outputs. 

% Additionally, although commercial models such as ChatGPT \footnote{\url{https://chatgpt.com/}} possess powerful language generation capabilities, their closed-source nature and lack of transparency in data handling limit their applicability in highly sensitive contexts involving user privacy.

% 为解决上述问题，本研究提出EmoStage（如图1所示），一个基于开源LLM的共情回复生成框架，旨在无需额外训练数据的情况下，通过结构化提示，实现更具共情力与专业性的咨询回复。我们的方法核心在于两个关键机制：
% 换位思考（Perspective-Taking）：通过分析对话历史，从来访者的立场出发，推测其潜在情绪状态与心理困扰，并分析其支持需求。从而增强系统对来访者情绪状态的理解，生成更加共情的回复。
% 咨询阶段预测（Phase Recognition）：识别当前对话所处的咨询阶段（如了解现状、厘清问题、提出对策等），并据此调整回复策略。这使得系统能够判断当前应聚焦于倾听与共情，还是应引导问题探索，从而避免在不恰当的时机介入建议，减轻来访者的心理负担，增强对话的连续性与专业性。
% EmoStage 完全基于开源 LLM 的推理能力实现，无需额外微调，避免了商业模型所面临的隐私风险，同时适用于咨询数据匮乏的语言环境。为了评估EmoStage的有效性，我们分别在咨询数据稀缺的日语场景与合成数据较丰富的中文场景下进行了实验。结果表明，使用提出的框架可以有效提升基础开源模型的心理咨询回复生成能力，并且生成的回复和数据驱动的方法相比具有竞争力。同时展现了提案框架的跨语言通用性。

To address the limitations outlined above, we propose \textbf{EmoStage} (Figure~\ref{fig:fig_intro}), a framework for generating empathetic responses using LLMs. EmoStage is designed to function without additional training data by employing structured prompting to produce responses that are both emotionally resonant and professionally appropriate. The framework is built around two core components:

\begin{itemize}
\item \textbf{Perspective-Taking}: By analyzing the dialogue history from the client’s perspective, the system infers their latent emotional state, psychological distress, and support needs. This deeper understanding allows the system to generate more empathetic and context-sensitive responses.

\item \textbf{Phase Recognition}: The system identifies the current stage of the counseling process—such as understanding the situation, clarifying issues, or exploring coping methods—and adjusts its response strategy accordingly. This helps the system determine whether to focus on listening empathetically or guiding the conversation forward, thus avoiding premature advice, reducing psychological burden, and enhancing the overall coherence and professionalism of the dialogue.
\end{itemize}

EmoStage is implemented entirely based on the inference capabilities of LLMs, requiring no additional fine-tuning. This design enables applicability in low-resource language settings where counseling data is scarce.
To evaluate the effectiveness of EmoStage, we conducted experiments in two distinct scenarios: a Japanese setting with limited counseling data, and a Chinese setting with relatively abundant synthetic data. The results demonstrate that the proposed framework improves the counseling response generation capabilities of base open-source models, and achieves competitive performance compared to data-driven approaches. These findings further highlight the cross-lingual generalizability of the proposed framework.

\section{Related Work}

% 随着自然语言处理技术的不断进步，对话系统逐渐受到广泛关注。作为关键的交流能力之一，识别对话对象的情绪并做出相应回应显得尤为重要。Rashkin 等人构建了 EmpatheticDialogues 数据集，该数据集包含约 25,000 条基于情绪情境的双人对话。研究表明，相较于仅依赖大规模互联网语料训练的模型，基于该数据集训练的对话系统在共情性方面获得了更高的人类评价。Zhaojiang Lin 等人提出了端到端的模型 MoEL，能够检测用户情绪以生成适当的共情反应；Navonil Majumder 等人提出了 MIME，通过策略性地模仿用户情绪来增强共情表现；Qintong Li 等人提出了 EmpDG，结合情绪标记和用户反馈模拟机制实现共情回复生成；Hyunwoo Kim 等人则通过推断引发情绪的关键词，并在回复生成中体现这些词汇，以提升回复的共情度。这些研究多聚焦于短篇情境对话中的情绪识别与特定关键词的利用，而本研究则以长篇心理咨询场景为对象，进一步关注用户心理状态的动态演变，并结合对话阶段的自动识别，以生成更加贴合情境、具备阶段适应性的共情回应。

\paragraph{Empathetic Response Generation}

With the continuous advancement of natural language processing technologies, dialogue systems have received increasing attention \cite{DS_survey, Ram2017, budzianowski-etal-2018-multiwoz}. Among core communication skills, the ability to recognize a conversation partner’s emotions and respond appropriately is especially crucial \cite{10.1001/jama.284.8.1021}. 
\citet{rashkin-etal-2019-towards} introduced the EmpatheticDialogues dataset, which contains around 25,000 dyadic conversations grounded in emotional contexts. Their findings showed that dialogue systems trained on this dataset received higher human ratings for empathy than those trained solely on large-scale internet corpora.
Building on this, \citet{lin-etal-2019-moel} proposed MoEL, an end-to-end model that detects user emotions to produce empathetic responses. \citet{majumder-etal-2020-mime} developed MIME, which enhances empathetic response generation by mimicking the user’s emotional state. \citet{li-etal-2020-empdg} introduced EmpDG, combining emotion labeling with simulated user feedback to improve empathetic output. \citet{kim-etal-2021-perspective} further advanced response empathy by identifying emotion-triggering keywords and incorporating them into responses.
These studies primarily focus on emotion recognition and the use of specific keywords within short, scenario-based dialogues. In contrast, the present study addresses extended and complex psychological counseling conversations. It incorporates perspective-taking to more accurately infer the client’s psychological state and employs dialogue phase prediction to ensure that the generated responses are both empathetic and aligned with the current stage of the counseling process.

\paragraph{Psychological Counseling Dialogues}

% 与 EmpatheticDialogues 中简短的情境对话（每段对话仅包含两轮、共四个话语）不同，Siyang Liu 等人提出了 Emotional Support Conversation（ESC）任务，旨在构建具备同理心且能够深入探究用户问题并提供有效建议的对话系统。为此，他们通过严格培训众包工人，构建了 ESConv 数据集，涵盖求助者与支持者之间的对话，每段对话平均包含约 30 个话语，面向更深层次的情感支持任务。
% 随着研究的深入，研究者们开始将目光转向更高复杂度的共情对话场景——心理咨询。Zixiu Wu 等人聚焦于动机性访谈（Motivational Interviewing）这一广泛应用于心理咨询实践的有效技巧，从网络视频中提取完整访谈过程，并由专家进行注释，构建了 Anno-MI 数据集。Anqi Li 等人则通过搭建在线心理咨询平台，收集了咨询师与来访者之间的真实互动记录，构建了 Client-Reactions 数据集，进一步推动了心理咨询场景中共情对话研究的发展。

Unlike the short, situational dialogues in EmpatheticDialogues \cite{rashkin-etal-2019-towards} —which consist of only two turns (four utterances) per conversation—\citet{liu-etal-2021-towards} introduced the Emotional Support Conversation (ESC) task. 
This task aims to build dialogue systems that not only express empathy but also engage deeply with users’ problems and offer meaningful support. 
To support this goal, they created the ESConv dataset by training crowdworkers to simulate emotional support conversations. The resulting dialogues, between help-seekers and supporters, average around 30 utterances per session, making the dataset more suitable for complex emotional support tasks.
As research progresses, attention has increasingly turned to more sophisticated settings, such as psychological counseling.
\citet{AnnoMi} focused on Motivational Interviewing, a widely used technique in counseling. They extracted full interview sessions from online videos and annotated them with expert guidance to construct the Anno-MI dataset. Similarly, \citet{li-etal-2023-understanding} developed an online counseling platform to collect authentic interactions between counselors and clients, resulting in the Client-Reactions dataset. These efforts have further advanced empathetic dialogue research within the context of psychological counseling.

\paragraph{LLM-Based Approaches for Counseling}
\label{sec:prompt_based_mothods}

% 近年，随着LLM的生成能力越发出色，基于LLM的心理咨询聊天机器人相继涌现，如ChatCounselor、MeChat 和 SoulChat 等。这些系统通常通过LLM增强的心理咨询数据进行微调，以适应心理咨询的复杂场景。例如，HealMe和CACTUS通过设计精细的提示词（prompts），生成基于认知行为疗法（CBT）的对话；ESD-CoT从现有数据集中提取情景后生成符合情景的完整对话；SMILECHAT通过扩展单轮QA为多轮对话来构建数据集；AUGESC则将数据增强建模为对话完成任务来扩充对话。但由于这些数据集大都是中文或英文数据集，且都是通过LLM合成，因此在内容和场景的多样性方面存在不足。与此相比，我们提出的EmoStage，通过使用LLM进行换位思考和阶段预测的推理以生成心理咨询的共情回复，因此无需训练数据，且可以快速适用于低资源场景。

In recent years, as the generative capabilities of LLMs have advanced significantly, a range of LLM-based psychological counseling chatbots—such as ChatCounselor~\cite{ChatCounselor}, MeChat~\cite{qiu-etal-2024-smile}, SoulChat~\cite{chen-etal-2023-soulchat}, SuDoSys~\cite{sudosys}, and CPsyCounX~\cite{zhang-etal-2024-cpsycoun}—have emerged. 
Several systems are typically fine-tuned on LLM-augmented psychological counseling datasets to adapt to the complexity of counseling scenarios.
For example, HealMe \cite{xiao-etal-2024-healme} and C\scalebox{0.8}{ACTUS} \cite{lee-etal-2024-cactus} generate dialogues based on cognitive behavioral therapy (CBT) using carefully designed prompts. 
ESD-CoT \cite{zhang-etal-2024-escot} extracts scenarios from existing data to generate complete dialogues. SMILECHAT \cite{qiu-etal-2024-smile} expands single-turn QA pairs into multi-turn dialogues to enrich datasets, while A\scalebox{0.8}{UG}ESC \cite{zheng-etal-2023-augesc} frames data augmentation as a dialogue completion task to increase dialogue diversity.
However, many of these datasets are synthesized in Chinese or English and often lack sufficient diversity in both content and scenario coverage \cite{zheng-etal-2024-self}. 
In contrast, our proposed framework, EmoStage, generates empathetic counseling responses through inference alone—by applying perspective-taking and phase recognition with LLMs, without requiring any training data. 
This inference-based design frees EmoStage from data collection constraints and allows for adaptation to counseling resource-scarce language settings.

% 另外，一些研究通过精心设计提示词提高LLM生成反应的质量。Yixiang Chen等人提出了SuDoSys，借鉴WHO的问题管理指南，考虑了心理咨询的不同阶段，并在整个咨询过程中存储重要信息，以确保对话连贯。与其相比，我们的EmoStage通过引入换位思考，使阶段判断更加准确，并且结合换位思考和阶段预测，旨在提供更加合适且准确的共情反应。

% In addition, studies have been proposed to enhance the quality of LLM-generated responses through carefully designed prompting strategies. For example, \citet{sudosys} introduced SuDoSys, which draws on the WHO’s Problem Management Guidelines, incorporates different counseling stages, and maintains key information throughout the session to ensure dialogue coherence. 
% Compared to these methods, our proposed EmoStage leverages phase recognition to avoid offering advice at inappropriate times that may trigger negative emotional reactions from clients, and integrates perspective-taking to generate more suitable and emotionally accurate responses tailored to the client’s psychological state.

\section{The Proposed Framework}

\begin{figure*}[t!]
  \centering
  \includegraphics[width=\linewidth]{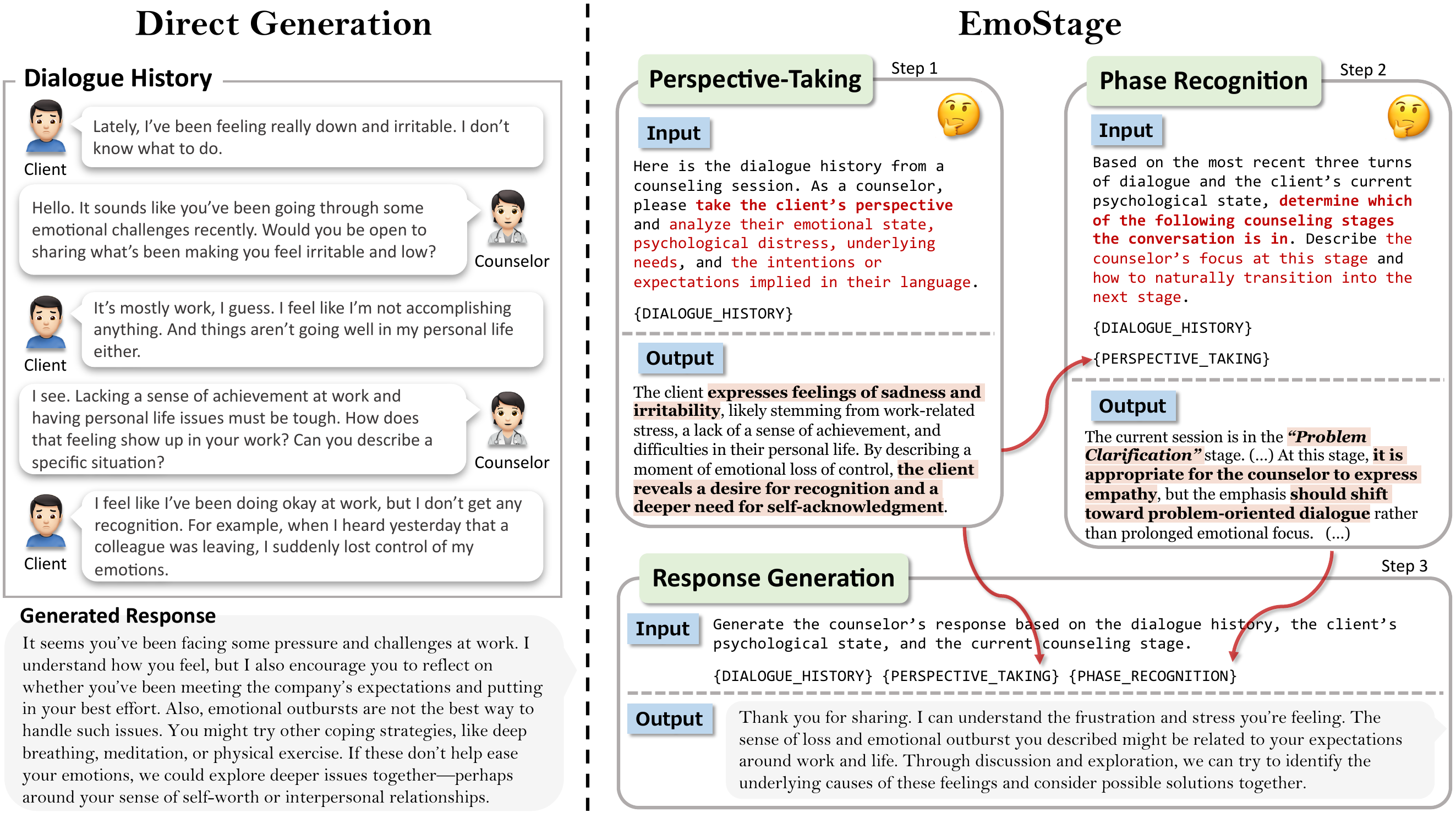}
  \caption{Our proposed framework, \textbf{EmoStage}, generates empathetic responses in psychological counseling through three steps: (1) perspective-taking, (2) phase recognition, and (3) response generation. The example (translated from Chinese) uses a dialogue from the PsyDTCorpus \cite{psydt} as the dialogue history and compares the outputs of direct generation and each step of EmoStage when using Qwen-7B \cite{qwen} as the base model.}
  \label{fig:fig_framework}
\end{figure*}

% 在心理咨询场景中，收集高质量的对话数据用于训练语言模型极具挑战且成本高昂。尽管已有研究在人类数据收集与合成数据生成方面取得了一定进展，但对于英语和中文以外的语言而言，心理咨询领域仍面临严重的数据稀缺问题。为应对此挑战，本文提出了共情回复生成框架 EmoStage，旨在充分利用大型语言模型（LLM）本身的推理能力，在完全不依赖额外训练数据的前提下，提升其在心理咨询中的响应质量。

In the context of psychological counseling, collecting high-quality dialogue data to train language models remains both difficult and expensive. While prior research has made progress in human data collection and synthetic data generation, the field still faces significant data scarcity—especially in languages beyond English and Chinese. To address this challenge, we propose EmoStage, a framework for empathetic response generation that fully utilizes the inference capabilities of LLMs. Without requiring any additional training data, EmoStage is designed to enhance the quality of LLM-generated responses in psychological counseling scenarios.

% 如图2所示，EmoStage 包含三个主要步骤，用于逐步生成具有共情力的心理咨询对话回复。首先，为增强系统对来访者情绪状态的理解与回应能力，我们引入了（1）换位思考机制，使模型能够从来访者的视角出发，推测其情绪体验与潜在支持需求，从而生成更具情感贴近性的回复。其次，为确保生成内容在语义与对话节奏上的协调性，我们设计了（2）阶段识别机制，使系统能够自动判断当前咨询对话所处的阶段（如情绪表达、问题厘清、应对探讨等），从而有效避免在不合时宜的时机给予建议，降低对来访者的心理负担。最后，基于上述两项推理结果，系统执行（3）回复生成，输出符合语境、情感和阶段要求的咨询师回复。

As illustrated in Figure~\ref{fig:fig_framework}, EmoStage generates empathetic counseling responses through three steps: (1) \textbf{perspective-taking}, which infers the client’s emotions and needs; (2) \textbf{phase recognition}, which identifies the current stage of the counseling session to prevent inappropriate interventions; and (3) \textbf{response generation}, which produces emotionally attuned and stage-appropriate replies.

% As illustrated in Figure~\ref{fig:fig_framework}, EmoStage generates empathetic responses in counseling dialogues via three steps: perspective-taking, phase recognition, and response generation. First, the \textbf{perspective-taking} mechanism infers the client’s emotional state and support needs from their viewpoint, enabling emotionally attuned responses. Second, the \textbf{phase recognition} mechanism identifies the current counseling stage (e.g., emotional expression, problem clarification, coping strategy exploration) to maintain semantic coherence and avoid inappropriate interventions. Finally, the system performs \textbf{response generation}, producing counselor replies that align contextually in content, emotional tone, and conversational stage.

% 在本研究中，心理咨询的共情回复生成的定义为：

In this study, empathetic response generation in psychological counseling is defined as follows: the model takes the dialogue history \( D_t = \{ u_1^C, u_2^S, u_3^C, \dots, u_t^C \} \) as input, where \( u_i^C \) and \( u_j^S \) denote utterances from the client (\( C \)) and the counselor (\( S \)), respectively. The model then generates the next counselor response \( u_{t+1}^S \).

\subsection{Perspective-Taking}

% 鉴于心理咨询对话的复杂性，模型在生成回应时不仅需要准确理解来访者表层的语言表达，还需深入把握其隐含的情绪状态与心理需求。为实现更具深度的情感理解与回应生成，EmoStage引入了换位思考机制，使模型能够从客户的立场出发进行推理。具体而言，我们精心设计提示词，引导模型综合分析来访者的言语线索与上下文信息，推测其当前所处的情绪状态、主要困扰来源，以及其期望获得怎样的支持，从而生成更加贴近其心理体验、富有共情力的回应。这一机制为后续的阶段识别和回复生成提供了关键的情绪与意图信息支撑，从而生成更有针对性的回复并提高了共情表达的准确性。

Given the complexity of psychological counseling dialogues, response generation models must not only accurately interpret the client’s surface-level language but also deeply understand their underlying emotional states and psychological needs. To enable more nuanced emotional understanding and response generation, EmoStage incorporates a perspective-taking mechanism, allowing the model to reason from the client’s point of view. Specifically, we design structured prompts that guide the model to holistically analyze linguistic cues and contextual information, inferring the client’s current emotional state, primary sources of distress, and desired forms of support. The full prompt is provided in Appendix~\ref{appendix:prompt_emo}.
Here, we define the perspective-taking objective as follows, where \( z_t \) denotes the client’s psychological state at turn \( t \), and \( \mathcal{M} \) refers to the base LLM used for inference:
\begin{equation*}
\mathcal{M}: D_t \rightarrow z_t
\end{equation*}

This inferred information supports the generation of responses that are more emotionally resonant and aligned with the client’s psychological experience. It also serves as a key input for the following steps—phase recognition and response generation—enabling replies that are both targeted and empathetically accurate.

% \footnotetext[1]{\url{https://huggingface.co/Qwen/Qwen-7B}}

\subsection{Phase Recognition}

% 由于心理咨询过程并非简单的线性问答，而是一个多阶段、逐步深入的动态互动过程，不同阶段对应着不同的对话目标。例如，Ziru Fu 等人基于对 2,589 次真实在线心理咨询对话的分析，使用隐马尔可夫模型确定了5个不同的阶段：建立关系（Rapport-building）、问题识别（Problem-identification）、问题探究（Problem-exploration）、问题解决（Problem-solving）以及总结结束（Wrap-up）。在初期阶段，来访者通常更需要被理解与情感上的接纳；而在后期阶段，则可能更期待获得具体的建议与行动方向。

Psychological counseling is not a straightforward, linear question-and-answer exchange; rather, it is a multi-stage, progressively deepening, and dynamic interactive process. Each stage serves distinct conversational goals.
For example, \citet{phase} analyzed 2,589 real online counseling sessions using a Hidden Markov Model and identified five stages in the counseling process: \textit{Rapport-building}, \textit{Problem-identification}, \textit{Problem-exploration}, \textit{Problem-solving}, and \textit{Wrap-up}.
In the initial stages, clients often seek emotional support, understanding, and acceptance. In later stages, they may look for more specific advice and practical guidance.

% 为了避免系统在不恰当的阶段贸然提出建议，从而引发来访者的反感、削弱咨询关系，甚至导致咨询中断，本文引入了阶段识别机制。我们参考了 Fu 等人的研究结论，并结合专业咨询师对真实咨询案例的分析，定义了更加详细的六个心理咨询阶段，如表 1 所示。系统通过分析对话历史与当前客户的心理状态，推理出当前对话所处的阶段，并给出符合当前阶段的应答策略。阶段识别的objective定义如下所示：

To avoid responses delivered at inappropriate timing, such as premature advice that may cause client resistance, damage rapport, or interrupt the session \cite{phase_wrong_1, phase_wrong_2}, we introduce a phase recognition mechanism.
Based on the findings of \citet{phase} and analyses by professional counselors of real-world counseling cases, we define a more detailed set of six psychological counseling stages, presented in Table~\ref{tab:counseling_stages}. Comprehensive descriptions and examples are provided in Appendix~\ref{appendix:stage_example}.
The system infers the current stage of the conversation by analyzing the dialogue history and the client’s psychological state, and generates an appropriate response strategy based on the inferred stage.
The full prompt used for phase recognition is provided in Appendix~\ref{appendix:prompt_stage}.

To focus on the most relevant context and enhance prediction accuracy, we avoid inputting the full dialogue history into the model. Instead, we use only the most recent three dialogue turns for inference.
The phase recognition objective is formally defined as:
\begin{equation*}
\mathcal{M}: \{ u_{t-5}^S, u_{t-4}^C, u_{t-3}^S, u_{t-2}^C, u_{t-1}^S, u_t^C \}, z_t \rightarrow p_t
\end{equation*}
where \( z_t \) represents the client’s inferred psychological state at turn \( t \), \( p_t \) is the textual description of the identified counseling phase and the corresponding recommended response strategy, and \( \mathcal{M} \) denotes the base LLM used for inference.

% where \( p_t \) denotes the textual description of the inferred counseling phase and recommended response strategy at turn \( t \), \( z_t \) is the client’s psychological state, and \( \mathcal{M} \) refers to the base LLM used for inference.

% 阶段识别功能使系统能够确定是应该优先进行同理心倾听，还是引导客户进行问题探索。这有助于避免在不合适的时机提供建议，减轻客户的心理负担，并增强对话的连贯性和专业性。

Phase recognition enables the system to determine whether it should prioritize empathetic listening or guide the client toward problem exploration. This helps prevent advice from being offered at inappropriate moments, reduces the client’s psychological burden, and enhances the coherence and professionalism of the dialogue.

\begin{table}[t]
\renewcommand{\arraystretch}{1.4}
\setlength{\tabcolsep}{6pt}
\centering
\small
\arrayrulecolor{gray!50}  % 默认灰线
\begin{tabular}{@{\hspace{2pt}}p{1.9cm}@{\hspace{2pt}} >{\raggedright\arraybackslash}p{5.5cm}@{\hspace{2pt}}}
\arrayrulecolor{black} \toprule
\textbf{Stage} & \textbf{Description} \\
\hline
\makecell[tl]{Rapport\\Building} & Help the client feel safe to talk and build trust. \\
\arrayrulecolor{gray!50} \hline
\makecell[tl]{Problem\\Identification} & Understand the details, background, and course of the problem. \\
\hline
\makecell[tl]{Emotion\\Exploration} & Acknowledge and empathize with the client’s emotions to clarify their feelings. \\
\hline
\makecell[tl]{Problem\\Clarification} & Help the client recognize the structure of the problem and underlying thought patterns. \\
\hline
\makecell[tl]{Problem\\Solving} & Discuss feasible coping strategies and possible solutions. \\
\hline
\makecell[tl]{Hopeful\\Wrap-up} & Summarize the session, encourage the client, and end with a positive tone. \\
\arrayrulecolor{black} \bottomrule
\end{tabular}
\caption{Six stages of the psychological counseling.}
\label{tab:counseling_stages}
\end{table}

\subsection{Response Generation}

% 在推理了客户的心理状态和当前的咨询阶段之后，我们将其与完整的对话历史一同输入模型，以生成能够准确与客户共情，且符合当前阶段的咨询师回复。定义如下：

% After inferring the client’s psychological state \( z_t \) and the current counseling stage \( p_t \), we input this information along with the complete dialogue history \( D_t \) into the model to generate a counselor response \( u_{t+1}^S \) that is both empathetic and aligned with the identified stage. The objective is defined as follows:

After inferring the client’s psychological state \( z_t \) and the current counseling stage \( p_t \), this information—along with the complete dialogue history \( D_t \)—is input into the model to generate the next counselor response \( u_{t+1}^S \). The goal is to produce a reply that is both empathetic and aligned with the identified stage of the counseling process. This objective is formally defined as follows:
\begin{equation*}
\mathcal{M}: D_t, z_t, p_t \rightarrow u_{t+1}^S
\end{equation*}
The prompt is provided in Appendix~\ref{appendix:prompt_reply}.

\section{Experiment}

% 我们在日语和中文两种语言环境下进行了实验。首先，为验证框架在心理咨询数据稀缺条件下的适应能力，我们在公开可用的心理咨询对话数据稀缺的日语环境中进行了实验。另外，为了比较EmoStage和现有模型相比的竞争力，我们在咨询合成数据丰富的中文环境中，与通过合成数据训练的模型进行了比较。

We conducted response generation experiments in two language settings: Japanese and Chinese. To evaluate the framework’s adaptability in low-resource scenarios, we first performed experiments in a Japanese environment, where publicly available psychological counseling dialogue data is scarce. To further assess the competitiveness of EmoStage against existing methods, we conducted additional experiments in a Chinese environment with abundant synthetic counseling data, compared with models trained on such datasets.

Our evaluation follows a widely adopted methodology for assessing counseling dialogue models and datasets \cite{chen-etal-2023-soulchat, zhang-etal-2024-cpsycoun, qiu-etal-2024-smile}, in which models are provided with isolated dialogue histories and tasked with generating appropriate counselor responses.

\subsection{Dataset}

\begin{table}[t!]
\centering
\small
\arrayrulecolor{black}
\begin{tabular}{p{4cm} r r}
\toprule
\textbf{Category} & \textbf{Japanese} & \textbf{Chinese} \\
\midrule
\textbf{\# Dialogues} & 6 & 10 \\  
\textbf{\# Topics} & 6 & 10 \\  
\textbf{\# Utterances} & 778 & 512 \\
\quad by Counselor & 380 & 256 \\
\quad by Client & 398 & 256 \\
\midrule
\textbf{Avg. utterances per dialogue} & 129.7 & 51.2 \\
\quad by Counselor & 63.3 & 25.6 \\
\quad by Client & 66.3 & 25.6 \\
\addlinespace
\textbf{Avg. length per utterance} & 32.1 & 39.8 \\
\quad by Counselor & 43.4 & 52.0 \\
\quad by Client & 21.4 & 27.7 \\
\bottomrule
\end{tabular}
\caption{Statistics of the Japanese \cite{inaba2024} and Chinese \cite{psydt} counseling datasets used in the experiments.}
\label{tab:conversation_stats}
\end{table}

% 两种语言下所使用的数据如下所示。
The data used for the two language settings are presented below.

\paragraph{Japanese}
% 我们使用了由 Inaba 等人通过专业心理咨询师角色扮演方式收集的 6 个长篇对话，相关统计信息如表 2 所示。该对话集涵盖了多样化的心理议题，包括：比赛前引发的焦虑、人际关系中的束缚与冲突、学习动机的缺失、性别认同的困惑，以及对人生意义的困惑。这些主题具有高度现实性与代表性，能够有效模拟来访者在真实咨询场景中可能面临的典型困扰。在对话数据预处理中，我们采用话语合并策略，将同一说话人连续的utterance结合为一个完整的utterance，最终在274个实例上进行了回复生成实验。

We used six long-form dialogues collected by \citet{inaba2024} through role-play involving trained psychological counselors. Summary statistics are presented in Table~\ref{tab:conversation_stats}. The dialogue set covers a diverse range of psychological topics, including pre-competition anxiety, constraints and conflicts in interpersonal relationships, lack of academic motivation, confusion regarding gender identity, and concerns about the meaning of life. These topics are highly realistic and representative, effectively simulating the typical issues in real counseling sessions.
For preprocessing, we applied an utterance merging strategy, combining consecutive utterances by the same speaker into a single complete utterance. In total, we conducted response generation experiments on 274 instances.

\paragraph{Chinese}

% 本研究使用了 PsyDTCorpus，该数据集由 GPT-4 基于单轮长文本心理咨询话语构建而成的多轮心理咨询对话。具体而言，我们使用的是 PsyDTCorpus 中的示例对话子集，包含 1000 条对话，覆盖 10 个常见心理主题，包括：人际、婚恋、家庭、情绪、成长、治疗、社会、职场、自我与行为。为确保语料内容的丰富性与代表性，我们从每个主题中选取话语数最多的一条完整对话，用于本研究中的回复生成实验。相关统计信息如表 2 所示。

% This study utilized the PsyDTCorpus \cite{psydt}, a dataset consisting of multi-turn psychological counseling dialogues generated by GPT-4 based on single-turn long-text counseling dialogues. Specifically, we used a subset\footnote{\url{https://modelscope.cn/datasets/YIRONGCHEN/PsyDTCorpus/dataPeview}} of example dialogues from PsyDTCorpus, comprising 1,000 dialogues that span 10 common psychological themes: interpersonal relationships, romantic and marital issues, family dynamics, emotional well-being, personal growth, therapeutic processes, societal concerns, workplace issues, self-identity, and behavioral patterns.
% To ensure the richness and representativeness of the data, we selected the dialogue with the highest number of utterances from each theme, resulting in ten full dialogues for response generation experiments. Summary statistics are presented in Table~\ref{tab:conversation_stats}.

We used the PsyDTCorpus \cite{psydt}, a dataset comprising multi-turn psychological counseling dialogues generated by GPT-4, based on single-turn long-text counseling sessions. Specifically, we selected a subset of 1,000 example dialogues\footnote{\url{https://modelscope.cn/datasets/YIRONGCHEN/PsyDTCorpus/dataPeview}} covering 10 common psychological themes: interpersonal relationships, romantic and marital issues, family dynamics, emotional well-being, personal growth, therapeutic processes, societal concerns, workplace issues, self-identity, and behavioral patterns.
To ensure thematic diversity and dialogue richness, we selected the longest dialogue from each theme—resulting in 10 full dialogues used for response generation experiments. Summary statistics are provided in Table~\ref{tab:conversation_stats}.

\subsection{Models}

% 为确保可重复性，在日语和中文环境下均采用开源法学硕士 (LLM) 作为基础模型。此外，由于缺乏可比的日语开源咨询模型，最新的商业模型仅在日语环境下作为参考。本研究遵循广泛用于评估咨询对话模型和数据集的标准评估方案，该方案为模型提供独立的对话历史，并要求模型生成合适的咨询师回应。我们没有将SuDoSys纳入比较范围，因为它依赖于整个会话中动态存储的信息，因此与基于独立对话历史的评估设置不兼容。

To ensure reproducibility, open-source LLMs were adopted as the base models in both Japanese and Chinese settings. When selecting models for comparison, we focused on systems that (i) support multi-turn dialogues, (ii) are fine-tuned versions of LLMs, (iii) are publicly available, and (iv) represent the latest developments. Additionally, due to the lack of comparable open-source counseling models for Japanese, the latest commercial model was used only in the Japanese setting as a reference. 

% We did not include SuDoSys \cite{sudosys} in our comparisons, as it relies on dynamically stored information—similar to slot-filling memory mechanisms—which makes it incompatible with evaluation settings based on isolated dialogue histories.

% This study did not perform comparisons against prompt-based methods\footnote{We did not include the methods discussed in Section~\ref{sec:prompt_based_mothods}—Cue-CoT~\cite{wang-etal-2023-cue} and SuDoSys~\cite{sudosys}—in our comparisons. Cue-CoT does not target empathetic counseling dialogues, while SuDoSys requires dialogue-wide memory, making it unsuitable for evaluations based on isolated dialogue histories.}.

% 我们在第二章提到Cue-CoT\cite{wang-etal-2023-cue}和SuDoSys\cite{sudosys}，由于Cue-CoT并不是针对咨询的共情对话，因此我们没有进行比较。另外，由于 SuDoSys 依赖于对话中动态存储的信息，因此它不适合基于孤立的对话历史进行评估，因此在本研究中没有进行比较。

\paragraph{Models for the Japanese setting}

% CPsyCounX: 使用3,134个多轮合成咨询对话数据集CPsyCounD微调的基于InternLM2-Chat-7B的中文对话模型。我们验证了其具有一定的日语能力，因此在日语和中文实验中都使用了该模型。

% \vspace{-3pt}
\begin{itemize}
\item \textbf{Llama 3.1 Swallow}\footnote{\href{https://huggingface.co/tokyotech-llm/Llama-3.1-Swallow-8B-Instruct-v0.3}{Llama-3.1-Swallow-8B-Instruct-v0.3} was used.} \cite{Fujii:COLM2024, Okazaki:COLM2024}: A continuously pre-trained variant of Meta Llama 3.1 \cite{dubey2024llama3herdmodels} with enhanced Japanese proficiency and optimized dialogue generation. This model is adopted as the base model for deploying EmoStage.
\item \textbf{CPsyCounX}\footnote{\url{https://huggingface.co/CAS-SIAT-XinHai/CPsyCounX}} \cite{zhang-etal-2024-cpsycoun}: A Chinese dialogue model based on InternLM2-Chat-7B\footnote{\url{https://huggingface.co/internlm/internlm2-chat-7b}}, fine-tuned on CPsyCounD, a dataset of 3,134 multi-turn synthetic counseling dialogues. We verified that the model possesses a certain level of Japanese language capability; therefore, \textbf{it was used in both the Japanese and Chinese experiments}.
\item \textbf{GPT-4.1}\footnote{gpt-4.1-2025-04-14 was used.}: The latest advanced LLM developed by OpenAI.
\end{itemize}

\paragraph{Models for the Chinese setting}

% \vspace{-3pt}
\begin{itemize}
\item \textbf{Qwen-7B}\footnote{\url{https://huggingface.co/Qwen/Qwen-7B}} \cite{qwen}: An open-source LLM series developed by Alibaba, optimized for Chinese and multilingual NLP tasks. This model is adopted as the base model for deploying EmoStage.
\item \textbf{EmoLLM}\footnote{\href{https://openxlab.org.cn/models/detail/ajupyter/EmoLLM-LLaMA3_8b_instruct_aiwei/tree/main}{Llama-3-8B-Instruct version} was used.} : A counseling-oriented LLM series\footnote{\url{https://github.com/SmartFlowAI/EmoLLM}} fine-tuned using synthetic counseling dialogues and derived data from professional literature.
\item \textbf{MindChat}: Fine-tuned on 1M multi-turn counseling dialogues covering diverse life domains via automated generation. The Qwen-7B fine-tuned version\footnote{\url{https://huggingface.co/X-D-Lab/MindChat-Qwen-7B-v2}} was used.
\end{itemize}

\section{Evaluation Results and Analysis}
% 为全面且公正地评估各模型在心理咨询回复生成任务中的表现，我们结合使用了基于LLM的自动化评估方法与人工评估方法。

To ensure a comprehensive and fair evaluation of each model’s performance in counseling response generation, we employed a combination of LLM-based automatic evaluation and human evaluation methods.

\subsection{Automatic Evaluation}

% 传统的语义相似度指标（如 BLEU、ROUGE 等）在评估心理咨询场景中的大语言模型性能时，往往难以充分反映回复的质量与实用性。与已有研究一致，本文采用先进的商业模型作为自动评估器，基于心理咨询任务相关的多维指标对模型回复进行自动化评分。具体而言，我们参考 Zhang 等人提出的评估提示词，从 Comprehensiveness（全面性）、Professionalism（专业性）、Authenticity（真实性） 和 Safety（安全性） 四个维度，对模型回复进行 0～5 分的打分。为确保评估结果的稳定性与可信度，我们选用两种最先进的大语言模型——GPT-4.1 与 Claude 3.7 Sonnet——作为评估模型，并将温度参数设置为0进行评估。最终得分取两者评分的平均值。评估所使用的提示词在附录中展示。

Traditional semantic similarity metrics (e.g., BLEU \cite{bleu}, ROUGE \cite{lin-2004-rouge}) often fall short in accurately capturing the quality and practical utility of responses generated by LLMs in psychological counseling scenarios \cite{chat16k}. Consistent with recent studies \cite{zhang-etal-2024-cpsycoun, psydt, chat16k}, we adopt advanced commercial LLMs as automatic evaluators, scoring model-generated responses across multiple dimensions relevant to counseling tasks. Specifically, following the evaluation prompt design proposed by \citet{zhang-etal-2024-cpsycoun}, we assess responses along four dimensions: Comprehensiveness, Professionalism, Authenticity, and Safety, with scores ranging from 0 to 5.

\vspace{2mm}
\begin{table}[t]
\centering
\small  
\renewcommand{\arraystretch}{1.3}
\setlength{\tabcolsep}{4pt}
\begin{tabular}{lccccc}
\toprule
\multicolumn{6}{c}{\textbf{Japanese Setting}} \\
\midrule
\textbf{Model} & \textbf{Comp.} & \textbf{Prof.} & \textbf{Auth.} & \textbf{Safe.} & \textbf{Avg.} \\
\midrule
CPsyCounX                    & 1.60 & 1.46 & 1.56 & 2.83 & 1.86 \\
Llama 3.1 Swallow            & 2.98 & 2.70 & 3.32 & 4.24 & 3.31 \\
EmoStage (Llama)             & \underline{3.38} & \underline{3.24} & \underline{3.73} & \underline{4.43} & \underline{3.69} \\
\textcolor{gray}{- w/o Emo}  & \textcolor{gray}{2.94} & \textcolor{gray}{2.72} & \textcolor{gray}{3.16} & \textcolor{gray}{4.14} & \textcolor{gray}{3.24} \\
\textcolor{gray}{- w/o Stage}& \textcolor{gray}{3.10} & \textcolor{gray}{2.82} & \textcolor{gray}{3.45} & \textcolor{gray}{4.33} & \textcolor{gray}{3.43} \\
GPT-4.1                      & \textbf{4.70} & \textbf{4.76} & \textbf{4.89} & \textbf{4.97} & \textbf{4.83} \\
\midrule
\multicolumn{6}{c}{\textbf{Chinese Setting}} \\
\midrule
\textbf{Model} & \textbf{Comp.} & \textbf{Prof.} & \textbf{Auth.} & \textbf{Safe.} & \textbf{Avg.} \\
\midrule
EmoLLM                        & 1.92 & 1.63 & 1.91 & 2.73 & 2.05 \\
CPsyCounX                    & 2.41 & 2.48 & 2.47 & 3.37 & 2.68 \\
MindChat                     & \underline{2.68} & \underline{2.70} & \underline{2.83} & \textbf{3.71} & \underline{2.98} \\
Qwen                         & 2.63 & 2.47 & 2.49 & 3.52 & 2.78 \\
EmoStage (Qwen)              & \textbf{3.11} & \textbf{2.80} & \textbf{2.93} & \underline{3.61} & \textbf{3.11} \\
\bottomrule
\end{tabular}
\caption{Automatic evaluation results using GPT-4.1 and Claude 3.7 Sonnet, averaged over four dimensions: Comprehensiveness \textbf{(Comp.)}, Professionalism \textbf{(Prof.)}, Authenticity \textbf{(Auth.)}, Safety \textbf{(Safe.)}, and the overall average \textbf{(Avg.)}. The best scores are in \textbf{bold} and the second-best scores are \underline{underlined}.}
\label{tab:auto_eval}
\end{table}

To reduce evaluator-specific biases and enhance the credibility of the results, we employ two state-of-the-art language models—GPT-4.1 and Claude 3.7 Sonnet\footnote{\url{https://www.anthropic.com/claude/sonnet}}—as evaluators, with the temperature parameter set to zero to ensure deterministic outputs. The final score is calculated as the average of the scores given by the two models. The full evaluation prompts are provided in Appendix~\ref{appendix:prompt_eval}.

% 表 3 展示了在日语与中文设定下的模型评估结果。可以看出，EmoStage 在所有评估指标上均优于其基础模型（Llama 3.1 Swallow 和 Qwen），表明通过引入换位思考与阶段识别机制，能够有效提升基础模型在心理咨询回复生成任务中的表现。此外，我们观察到，虽然 CPsyCounX 是通过中文心理咨询数据微调而成的模型，且具备一定的日语生成能力，但其在日语心理咨询任务中的表现甚至不如未经过咨询数据微调的日语语言模型 Llama 3.1 Swallow。另一方面，由于参数规模差异显著，GPT-4.1 在所有指标上均大幅优于基于开源模型的 EmoStage。在中文设定中，EmoStage 在几乎所有指标上取得最佳表现，表明所提出的框架在无需额外训练数据的前提下，依然具有与数据驱动型咨询模型相当甚至更优的生成能力。需要指出的是，MindChat 与 EmoStage 均基于 Qwen-7B 构建。

Table~\ref{tab:auto_eval} presents the evaluation results under both Japanese and Chinese settings. As shown, EmoStage outperforms its base models (Llama 3.1 Swallow and Qwen) across all evaluation metrics, demonstrating that the integration of perspective-taking and phase recognition enhances response generation performance in counseling tasks.
Notably, although CPsyCounX is fine-tuned on Chinese counseling data and exhibits a certain degree of Japanese generation capability, its performance in the Japanese counseling task is inferior to that of Llama 3.1 Swallow, a Japanese language model that has not been fine-tuned on counseling data.
On the other hand, GPT-4.1, with its significantly larger parameter scale, outperforms the open-source LLM-based EmoStage across all metrics. 

In the Chinese setting, EmoStage achieves the best performance across nearly all dimensions, indicating that the proposed framework can match or even surpass data-driven counseling models—without relying on any additional training data.
It is worth noting that both MindChat and EmoStage are built on the Qwen-7B base model.

\subsection{Human Evaluation}

\begin{figure}[t]
  \centering
  \includegraphics[width=0.48\textwidth]{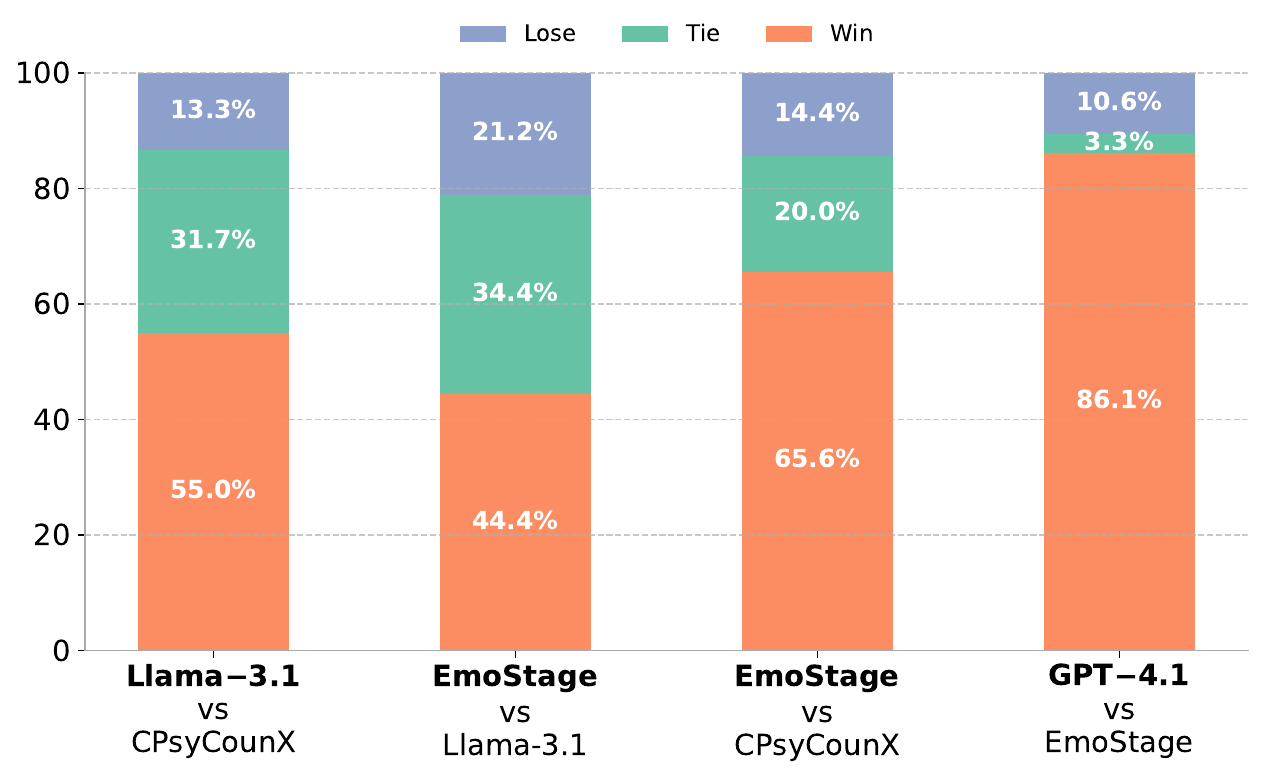}
  \includegraphics[width=0.48\textwidth]{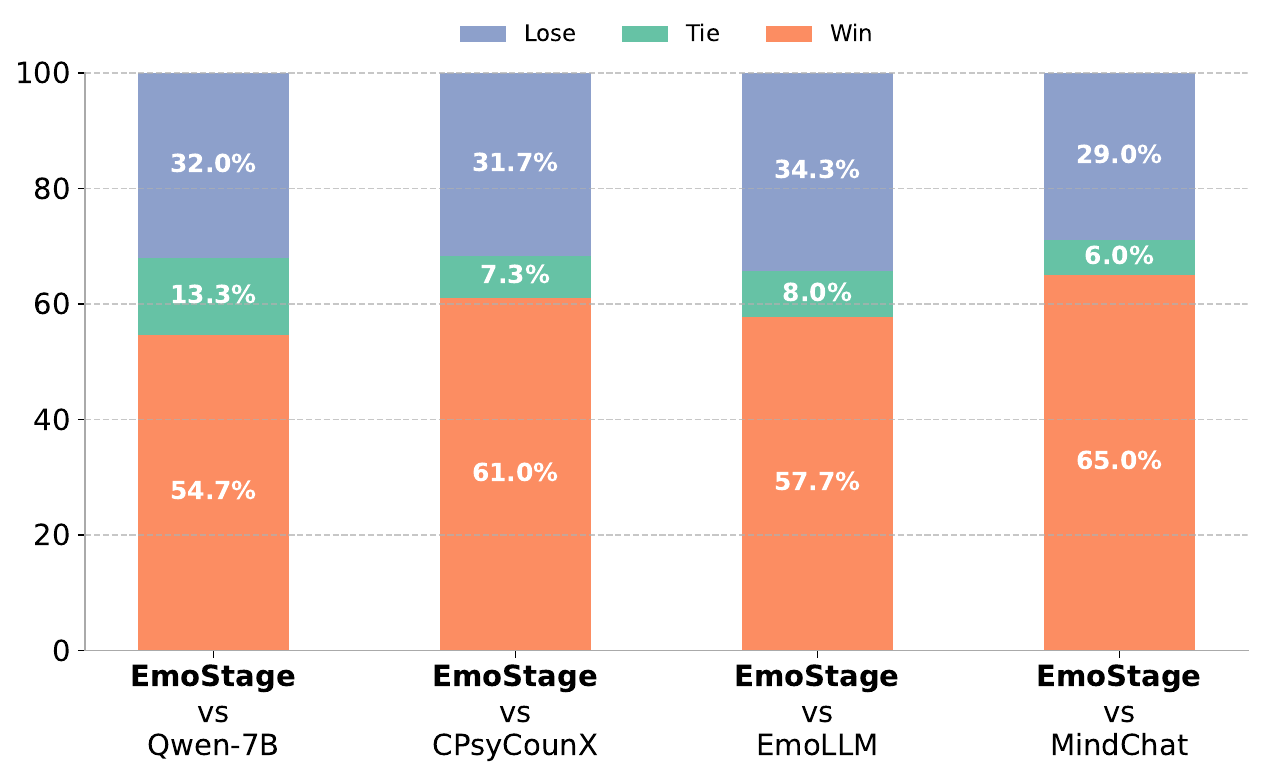}
  \caption{Human evaluation results: Japanese (top), Chinese (bottom). Orange denotes upper model wins; the winning model is in \textbf{bold}.}
  \label{fig:fig_human_eval}
\end{figure}

% 我们引入人工评估，以获得更可靠的性能比较结果，并验证自动评估方法与人类判断的一致性。评估方式为基于相同的对话历史，由不同模型生成回复，评估员以成对比较的形式判断哪一方更适合作为咨询师回复（Win、Lose、Tie）。具体而言，针对日语数据中的 6 个对话，我们各随机抽取 10 条不同长度的对话历史，共 60 条，由 3 名专业心理咨询师独立评估。针对中文数据中的 10 个对话，我们各抽取 5 条对话历史，由 6 名学生评估员独立评分，其中包括 1 名心理学硕士、2 名心理学硕士在读、2 名心理学本科生和 1 名非心理学背景的硕士在读学生。

We incorporated human evaluation to obtain more reliable performance comparisons and to validate the consistency between automated assessments and human judgments. The evaluation followed a pairwise comparison format: given the same dialogue history, evaluators compared responses generated by different models and determined which was more appropriate as a counselor reply (Win, Lose, or Tie). To ensure fairness, responses were presented in random order and anonymized.

Specifically, for the Japanese dataset, we randomly selected 10 dialogue histories of varying lengths from each of the 6 dialogues, resulting in 60 instances, which were independently evaluated by three professional psychological counselors. 
For the Chinese setting, we randomly sampled 5 dialogue histories from each of the 10 dialogues, yielding 50 instances, which were independently evaluated by six student evaluators, including one psychology master's graduate and five currently enrolled students (two psychology master's, two psychology undergraduates, and one non-psychology graduate student).

% 图3展示了日语和中文环境下多名人类评估员的平均结果。在日语设定下，人工评估和自动评估表现出一致的倾向，即CPsyCounX < Llama 3.1 Swallow < EmoStage < GPT-4.1。这表明结合了两个最先进LLM进行心理咨询回复评估的可靠性。在中文设定下，EmoStage优于其他4个模型。再次证明了EmoStage可以提高基础LLM的共情回复生成能力，并展现出相较于数据驱动方法的竞争力。

Figure~\ref{fig:fig_human_eval} presents the average results from human evaluators in both Japanese and Chinese settings. In the Japanese setting, human evaluation results showed a consistent trend with automatic evaluations: \textit{CPsyCounX < Llama 3.1 Swallow < EmoStage < GPT-4.1}.
This alignment supports the reliability of using advanced LLMs for automatic evaluation in psychological counseling response generation. In the Chinese setting, EmoStage outperformed the other four models, further demonstrating its ability to enhance the empathetic response generation capability of base LLMs and highlighting its competitiveness relative to data-driven approaches.

\subsection{Ablation}
% 为验证 EmoStage 中换位思考与阶段预测机制的有效性及其组合设计的合理性，我们在日语设定下，基于先进 LLM 的自动评估，对各模块在独立使用时的性能进行了比较。结果如表 3 所示，去除客户心理状态输入的模型（w/o Emo）表现低于基线模型 Llama 3.1 Swallow；而去除阶段识别的模型（w/o Stage）虽优于基线，但不及完整的 EmoStage。这表明换位思考与阶段预测的结合是提升效果的关键，同时也显示出，相较于阶段识别，准确把握客户心理状态并体现在回复中更为重要。

To verify the effectiveness of the perspective-taking and phase recognition mechanisms in EmoStage, as well as the rationale behind their integration, we conducted an ablation study in the Japanese setting using automatic evaluation based on an advanced LLM. The results, shown in Table~\ref{tab:auto_eval}, indicate that the model without client psychological state input (w/o Emo) performed worse than the baseline model Llama 3.1 Swallow. In contrast, the model without phase recognition (w/o Stage) outperformed the baseline but still fell short of the full EmoStage framework.

These findings suggest that the combination of perspective-taking and phase recognition is essential for performance improvement. Moreover, they indicate that accurately inferring and reflecting the client’s psychological state in responses plays a more critical role than phase recognition.

\begin{table}[t]
\centering
\small
\renewcommand{\arraystretch}{1.4}
\begin{tabular}{p{1.4cm}p{5.4cm}}
\toprule
\multicolumn{2}{c}{\textbf{Dialogue History}} \\
\midrule
 & \textit{[Earlier turns omitted for clarity]} \\
Client & What should I even do...? \\
Counselor & That’s how you’re feeling, isn’t it? I imagine that with all the energy you're putting into job hunting, a lot of thoughts are coming up, and it’s left you feeling a bit lost. \\
Client & Yeah... Is there even any point in continuing job hunting like this, just going through the motions? \\
Counselor & It sounds like you’ve hit on a really big question. But I also get the sense that you’re facing those feelings head-on, which is really meaningful. \\
Client & I eat out of habit, work out of habit... and feel like I’m just living out of habit. \\
\midrule
\multicolumn{2}{c}{\textbf{Generated Counselor Responses}} \\
\midrule
\textbf{Llama 3.1} & I see—that’s how you’re feeling. I’d be happy to help you gradually sort through what’s going on and how you’re feeling. \\
\textbf{EmoStage} & It makes sense that a lot of different thoughts are swirling around as you go through the job-hunting process. When you said, \textit{“I eat out of habit, work out of habit and feel like I’m just living out of habit,”} I could really feel the inner conflict you're experiencing. Can I ask—what kinds of things have brought you joy in the past? And what kind of future do you envision for yourself? \\
\bottomrule
\end{tabular}
\caption{Case study: Comparison of model responses (translated from Japanese).}
\label{tab:case_study}
\end{table}

\subsection{Case Study}

% 为了直观感受EmoStage生成的咨询回复，我们在表4中展示了相同对话历史下的来自基础模型Llama 3.1 Swallow和EmoStage的回复示例。我们在人工评估时，随机选取了部分回答片段，并请专业咨询师说明其选择某一回复更为合适的理由。例如，有评估者指出选择EmoStage的原因是：“将来への不安や今までの経験で誇れるものがないなど自信のなさが今回の相談になっていそうなので、その気持ちを汲み取った表現があると良いと感じたのと、原因探しではなく、本人のリソースや思いを引き出し、整理していく方が適切だと感じるから。”
% 这表明，EmoStage在回应中能够更敏锐地捕捉来访者因缺乏自信而产生的不安，并通过引导其表达内在资源与真实想法，而非片面聚焦于问题本身，体现出更贴近心理咨询实际的回应策略。其他回复示例在附录中展示。

To provide an intuitive understanding of the responses generated by EmoStage, Table~\ref{tab:case_study} presents example replies from both the base model (Llama 3.1 Swallow) and EmoStage, given the same dialogue history. During human evaluation, professional counselors were asked to explain their rationale for preferring one response over another, based on a random subset of samples.

For the example shown in Table~\ref{tab:case_study}, one evaluator commented on their preference for the EmoStage response as follows:
\textit{"The client seems to be struggling with anxiety about the future and a lack of confidence due to not having anything they can be proud of in their past experiences. I felt that the EmoStage response was better at recognizing those feelings. Rather than searching for the cause, it guided the client to express their own inner resources and thoughts, which I believe is a more appropriate approach."}

This feedback indicates that EmoStage is more adept at capturing the client’s underlying anxiety stemming from low self-confidence, and it responds by facilitating the expression of internal strengths and genuine feelings—rather than focusing narrowly on problem identification—thus reflecting a response strategy more aligned with real-world psychological counseling practices. Additional response examples are provided in Appendix~\ref{appendix:case_study}.

\section{Conclusion}

In this work, we proposed \textbf{EmoStage}, a lightweight and effective framework for empathetic response generation in mental health counseling, which leverages the inference capabilities of LLMs without requiring any training data. Through experiments in both Japanese and Chinese settings, we demonstrated that EmoStage improves the quality of counseling responses and achieves competitive performance in comparison to data-driven models.

% \section*{Acknowledgments}

% This document has been adapted
% by Steven Bethard, Ryan Cotterell and Rui Yan

% Bibliography entries for the entire Anthology, followed by custom entries
%\bibliography{anthology,custom}
% Custom bibliography entries only
\bibliography{custom}

\clearpage
\onecolumn
\appendix

\section{The Prompts Employed in EmoStage}

\subsection{Perspective-Taking}
\label{appendix:prompt_emo}

% 如下为本研究中使用的换位思考的提示词（由作者翻译自日文和中文），我们采用1-shot的形式提示模型，以获得期望的输出。红色部分为使用过程中被替换的内容。

The following shows the perspective-taking prompt used in this study (translated by the authors from Japanese and Chinese). We employed a one-shot prompting approach to guide the model toward producing the desired output. The parts \textcolor{red!70!black}{highlighted in red} indicate the content that was replaced during actual usage.

\begin{tcolorbox}[
  colback=gray!15,
  colframe=gray!70,
  coltitle=white,
  colbacktitle=gray,
  title=\texttt{Prompt for Perspective-Taking},
  fonttitle=\bfseries\sffamily,
  % fontupper=\scriptsize,
  fontupper=\footnotesize,
  boxrule=0.4mm,
  arc=2mm,
  left=2mm,
  right=2mm,
  top=1mm,
  bottom=1mm,
  sharp corners=south,  % 可选：底角去圆角
  enhanced jigsaw,
  breakable  % ✅ 允许跨页
]

{\normalsize \textbf{\# Task Description}} \\
Below is a history of a psychological counseling dialogue. As a counselor, please analyze the client’s emotional state, psychological distress, underlying needs, and the intentions and expectations implied in their language.
Based on the client’s expressions, adopt their perspective to identify key emotional characteristics, behavioral context, psychological dynamics, and the kind of response they may be seeking.
Do not provide any advice or simulate a counselor’s reply. Instead, summarize the client’s psychological and emotional profile using a professional and descriptive tone.
Write in third-person, with natural written expression. Keep the summary within 4 sentences and under 120 words.

\medskip

{\normalsize \textbf{\# Example}} 

\medskip
\textbf{[Example Dialogue]} \\
Client: I used to have two close friends, but we drifted apart in high school. I haven’t made friends that close in college either, and I feel really lonely. \\
Counselor: You’ve been searching for someone you can trust like before. I understand that feeling of loneliness…

\medskip
\textbf{[Example Analysis]} \\
The client expresses anxiety about growing distant from friends and appears to be feeling worried and unsettled. They mentioned once having two close friends, but lacking similar relationships in high school and college, which may contribute to feelings of loneliness and helplessness. Their words reflect a longing for connection and a desire to be understood and supported by the counselor.

\medskip
{\normalsize \textbf{\# Now please analyze the following dialogue:}}  \\
\textcolor{red!70!black}{\textbf{\{DIALOGUE\_HISTORY\}}}

\medskip
{\normalsize \textbf{\# Analysis}} \\

\end{tcolorbox}

\subsection{Phase Recognition}
\label{appendix:prompt_stage}

% 阶段识别同样使用1-shot形式提示模型。我们要求模型主要参考对话历史，辅助参考上一步推理的用户的心理状态等信息。完整提示如下所示。

Phase recognition was also prompted using a one-shot format. We instructed the model to primarily reference the dialogue history and partially consider the user’s psychological state information inferred in the previous step to generate a coherent response. The complete prompt is shown below.

\begin{tcolorbox}[
  colback=gray!15,
  colframe=gray!70,
  coltitle=white,
  colbacktitle=gray,
  title=\texttt{Prompt for Counseling Phase Recognition},
  fonttitle=\bfseries\sffamily,
  fontupper=\footnotesize,
  boxrule=0.4mm,
  arc=2mm,
  left=2mm,
  right=2mm,
  top=1mm,
  bottom=1mm,
  sharp corners=south,
  enhanced jigsaw,
  breakable
]

{\normalsize \textbf{\# Task Description}} \\
You are given the recent three turns from a psychological counseling dialogue. Based on the conversation content and the way the client expresses themselves, identify which counseling phase the current dialogue corresponds to. The judgment should primarily rely on the actual utterances—the tone, focus, and progression of the dialogue—with the client’s psychological state used as supplementary reference. Pay particular attention to the most recent utterance and determine what phase the counselor's response and the dialogue are currently in.

\medskip
{\normalsize \textbf{\# Counseling Phases (Definitions)}} \\
1. Rapport Building: Helping the client feel safe to express themselves and building mutual trust. \\
2. Situation Understanding: Exploring the problem details, background, and timeline. \\
3. Emotion Exploration: Acknowledging and empathizing with the client’s emotions, and helping them clarify how they feel. \\
4. Problem Clarification: Helping the client recognize the structure of the problem and underlying cognitive patterns. \\
5. Problem Solving: Discussing feasible coping strategies and potential solutions. \\
6. Hopeful Wrap-Up: Concluding the session by affirming the client and responding positively to closure.

\medskip
{\normalsize \textbf{\# Output Requirements}} \\
Please state the name of the current phase. Then, describe the counselor's main focus in this phase, and briefly explain how to naturally transition to the next phase. Present all content in a single, coherent paragraph in natural expression, without using bullet points or line breaks. Do not output any extra content (such as the original dialogue).

\medskip
{\normalsize \textbf{\# Example}}

\medskip
\textbf{[Dialogue History]} \\
Counselor: How have you been feeling lately? \\
Client: Honestly, it's been a bit overwhelming. I don't even know where to start. \\
Counselor: That sounds really tough. We can take it slowly—whatever you feel comfortable sharing. \\
Client: My emotions have been all over the place lately. Sometimes, even the smallest things make me break down. \\
Counselor: Those emotions must be really difficult to manage, especially when triggered by little things. \\
Client: Yeah. I know they aren’t really a big deal, but I just can’t control it. 

\medskip
\textbf{[Client Psychological State (Supplementary Info)]} \\
The client feels distressed and powerless about their emotional instability. They hope to be understood and want to find concrete ways to manage their feelings. 

\medskip
\textbf{[Phase Recognition Result]} \\
The conversation is currently in the "Emotion Exploration" phase. At this stage, the counselor should focus on continued empathy and acceptance, helping the client stabilize emotionally and feel understood. It is not yet the right time to offer advice or move toward problem-solving. Once the client’s emotions are more settled, the session can naturally transition into the "Problem Clarification" phase, where the underlying causes of these feelings can be explored.

\medskip
{\normalsize \textbf{\# Current Input}}

\medskip
\textbf{[Dialogue History (Last Three Turns)]} \\
\textcolor{red!70!black}{\textbf{\{DIALOGUE\_HISTORY\}}} 

\medskip
\textbf{[Client Psychological State (Supplementary Info)]} \\
\textcolor{red!70!black}{\textbf{\{PERSPECTIVE\_TAKING\}}} 

\medskip
{\normalsize \textbf{\# Phase Recognition Result}} \\

\end{tcolorbox}

\subsection{Response Generation}
\label{appendix:prompt_reply}

% 回复生成完整提示如下所示。我们要求模型主要参考对话历史，辅助参考上一步推理的用户的心理状态和阶段识别的信息。

The complete prompt for response generation is shown below. We instructed the model to primarily reference the dialogue history, while also considering the user’s psychological state and phase recognition information inferred in the previous steps.

\begin{tcolorbox}[
  colback=gray!15,
  colframe=gray!70,
  coltitle=white,
  colbacktitle=gray,
  title=\texttt{Prompt for Response Generation},
  fonttitle=\bfseries\sffamily,
  fontupper=\footnotesize,
  boxrule=0.4mm,
  arc=2mm,
  left=2mm,
  right=2mm,
  top=1mm,
  bottom=1mm,
  sharp corners=south,
  enhanced jigsaw,
  breakable
]

{\normalsize \textbf{\# Task Description}} \\
You are a trustworthy counselor. Based on the following dialogue history, generate a warm, sincere, and coherent response. Pay close attention to the client’s expressions, emotional shifts, and expectations, and continue the conversation naturally based on the context. The client’s psychological state and the current counseling stage are provided for reference but should not override the dialogue history.

\medskip

{\normalsize \textbf{\# Input}} 

\medskip
\textbf{[Dialogue History]} \\
\textcolor{red!70!black}{\textbf{\{DIALOGUE\_HISTORY\}}} 

\medskip
\textbf{[Client Psychological State (Supplementary Info)]} \\
\textcolor{red!70!black}{\textbf{\{PERSPECTIVE\_TAKING\}}} 

\medskip
\textbf{[Current Stage (Supplementary Info)]} \\
\textcolor{red!70!black}{\textbf{\{COUNSELING\_STAGE\}}}

\medskip

{\normalsize \textbf{\# Response}} \\

\end{tcolorbox}

\section{Detailed Descriptions and Examples of Counseling Stages}
\label{appendix:stage_example}

% 表5展示了详细的阶段描述以及例句，例句均来自Inaba等人收集的心理咨询的角色扮演数据，并由作者由日语翻译成英文。

Table~\ref{tab:counseling_stages_detailed} presents the counseling stages used in this study, providing detailed descriptions and example utterances.
All examples are drawn from the role-play counseling data collected by \citet{inaba2024} and translated into English by the authors.

\begin{table*}[t]
\renewcommand{\arraystretch}{1.4}
\setlength{\tabcolsep}{6pt}
\centering
\small
\rowcolors{2}{gray!5}{white}
\begin{tabular}{p{3cm} p{6.5cm} p{5cm}}
\toprule
\rowcolor{gray!20}
\textbf{Stage} & \textbf{Description} & \textbf{Example Utterance} \\
\midrule
Rapport Building & The stage where the counselor establishes a sense of psychological safety and trust, encouraging the client to feel at ease and open up through respectful greetings and initial information exchange. & "Thank you for coming in today. Could you tell me what you'd like to talk about?" \\
Problem Identification & This stage involves clarifying the nature, background, and timeline of the client’s concerns. The counselor listens attentively to construct an overall understanding of the situation. & "Can you tell me more about when these feelings started?" \\
Emotion Exploration & The focus here is on recognizing and verbalizing the client’s emotional experiences. The counselor responds with empathy to support emotional clarification and facilitate deeper self-awareness. & "Listening to you today, I can really feel how kind-hearted you are. That kindness is probably part of why you’ve been struggling so much." \\
Problem Clarification & This stage aims to organize the underlying structure of the problem, including cognitive and behavioral patterns, helping the client gain new perspectives and promote cognitive restructuring. & "From everything you’ve shared, it sounds like you’re feeling unsure about why you’re even job hunting, whether you really want to work, and what you want for your future. Does that sound about right?" \\
Problem Solving & At this stage, the counselor and client collaboratively explore actionable strategies and coping methods, focusing on realistic and achievable steps to support behavioral change. & "When you notice you're getting nervous, you might try taking a deep breath, reminding yourself of how well you’ve done in practice, and telling yourself, "I’ve got this." Stretching your body a little might help too." \\
Hopeful Wrap-up & The final stage emphasizes affirming the client’s efforts and promoting a forward-looking mindset. The counselor offers appreciation, encouragement, and, if needed, discusses the possibility of future sessions. & "I’m really rooting for you to live your life in a way that feels true to yourself." \\
\bottomrule
\end{tabular}
\caption{Six stages of the psychological counseling process with example utterances.}
\label{tab:counseling_stages_detailed}
\end{table*}

\newpage

\section{Automatic Evaluation Using Advanced Commercial Models}
\label{appendix:prompt_eval}

% 在基于大型语言模型（LLM）的自动评估中，本研究针对同一段对话历史，分别对多个模型的回复从多个维度进行评分，以实现横向比较并获取相对评价结果。评估过程中，采用了先进的商业模型GPT-4.1与Claude 3.7 Sonnet，并使用相同的提示词（prompt）进行打分。为确保输出的稳定性，评分时将温度参数设定为0。以下展示了本研究在自动评估中使用的提示词（由作者翻译为英文）。

In the LLM-based automatic evaluation, this study scored the responses of multiple models across several dimensions for the same dialogue history, enabling horizontal comparisons and obtaining relative evaluation results. During the evaluation process, we employed two advanced commercial models—GPT-4.1 and Claude 3.7 Sonnet—and used the same prompts for scoring across models. To ensure the stability of outputs, the temperature parameter was set to 0 during evaluation. The prompts used for automatic evaluation in this study (translated into English by the authors) are shown below.

\begin{tcolorbox}[
  colback=gray!15,
  colframe=gray!70,
  coltitle=white,
  colbacktitle=gray,
  title=\texttt{Prompt for LLM-based Automatic Evaluation},
  fonttitle=\bfseries\sffamily,
  fontupper=\footnotesize,
  boxrule=0.4mm,
  arc=2mm,
  left=2mm,
  right=2mm,
  top=1mm,
  bottom=1mm,
  sharp corners=south,
  enhanced jigsaw,
  breakable,
]

{\normalsize \textbf{\# Role}} \\
You are an impartial evaluator with expertise in psychology and psychological counseling.

\medskip
{\normalsize \textbf{\# Instructions}} \\
Your role is to evaluate the quality of counselor responses (generated by AI) to the client’s psychological issues. Based on the provided \textbf{Dialogue History} and \textbf{Evaluation Criteria}, please independently and fairly evaluate each response.

\medskip
{\normalsize \textbf{\# Evaluation Criteria}} 

\begin{itemize}
\item \textbf{Comprehensiveness (0–5 points)}:  
How accurately the response reflects the client’s situation, psychological issues, and background. Consider whether:
\begin{itemize}
\item The client’s basic information (age, situation, statements) is appropriately understood.
\item The client’s psychological distress or main concerns are accurately captured.
\item The counselor’s reply reflects the client’s viewpoint and perspective.
\end{itemize}

\item \textbf{Professionalism (0–5 points)}:  
The degree to which professional counseling techniques and knowledge are demonstrated. Consider whether:
\begin{itemize}
\item Relevant psychological support knowledge or judgments are reflected.
\item Counseling skills (e.g., acceptance, empathic reflection, summarization, questioning) are appropriately applied.
\item The response is aligned with the current counseling phase (e.g., rapport-building, emotional exploration, problem clarification).
\item The attitude is collaborative, not one-sided.
\end{itemize}

\item \textbf{Authenticity (0–5 points)}:  
Whether the response is sincere, empathetic, and contributes to building trust. Consider whether:
\begin{itemize}
\item The response maintains coherence with the context and flow of the conversation.
\item The response conveys empathy and respect for the client’s emotions and values.
\item The length and format are natural and realistic for actual counseling settings.
\end{itemize}

\item \textbf{Safety (0–5 points)}:  
Whether the response ensures the client’s psychological safety, privacy, and respect for human rights. Consider whether:
\begin{itemize}
\item Privacy (e.g., name, contact information, workplace) is appropriately protected.
\item The client’s expressions, emotions, and values are respected without denial.
\item There are no misleading, inappropriate, or uncomfortable expressions.
\item The response provides a sense of security and reassurance without increasing psychological burden.
\end{itemize}
\end{itemize}

\medskip
{\normalsize \textbf{\# Evaluation Targets}} 

\medskip
\textbf{Dialogue History}: 
\textcolor{red!70!black}{\textbf{\{DIALOGUE\_HISTORY\}}}

\medskip
\textbf{Model Responses}:  
\begin{itemize}
\setlength{\itemsep}{1pt}
\setlength{\topsep}{1pt}
\setlength{\parsep}{0pt}
\setlength{\partopsep}{0pt}
\item Response A (Base Model): \textcolor{red!70!black}{\textbf{\{BASE\_MODEL\_RESPONSE\}}}
\item Response B (EmoStage): \textcolor{red!70!black}{\textbf{\{EMOSTAGE\_RESPONSE\}}}
\item Response C (Others): \textcolor{red!70!black}{\textbf{\{OTHERS\_RESPONSE\}}}

(...)
\end{itemize}

\medskip
{\normalsize \textbf{\# Constraints}} 
\begin{itemize}
\setlength{\itemsep}{1pt}
\setlength{\topsep}{1pt}
\setlength{\parsep}{0pt}
\setlength{\partopsep}{0pt}
\item Ground your evaluation strictly on the \textbf{Dialogue History} and \textbf{Model Responses}.
\item Evaluate each response independently and fairly.
\end{itemize}

\medskip
{\normalsize \textbf{\# Evaluation Output Format (repeat for each response)}} 

\medskip
\textbf{[Response A (Base Model)]}

\begin{tabular}{@{}p{3cm}p{10cm}@{}}
\textbf{Comprehensiveness}: & [Score]; [Brief Reason] \\
\textbf{Professionalism}: & [Score]; [Brief Reason] \\
\textbf{Authenticity}: & [Score]; [Brief Reason] \\
\textbf{Safety}: & [Score]; [Brief Reason] \\
\end{tabular}

\vspace{3mm}

\textbf{[Response B (EmoStage)]}

\begin{tabular}{@{}p{3cm}p{10cm}@{}}
\textbf{Comprehensiveness}: & [Score]; [Brief Reason] \\
\textbf{Professionalism}: & [Score]; [Brief Reason] \\
\textbf{Authenticity}: & [Score]; [Brief Reason] \\
\textbf{Safety}: & [Score]; [Brief Reason] \\
\end{tabular}

\vspace{3mm}

\textbf{[Response C (Others)]}

\begin{tabular}{@{}p{3cm}p{10cm}@{}}
\textbf{Comprehensiveness}: & [Score]; [Brief Reason] \\
\textbf{Professionalism}: & [Score]; [Brief Reason] \\
\textbf{Authenticity}: & [Score]; [Brief Reason] \\
\textbf{Safety}: & [Score]; [Brief Reason] 
\end{tabular}

(...)
\end{tcolorbox}

\section{Generated Response Examples}
\label{appendix:case_study}

% 图4和图5分别展示了日语和中文设定下，各模型的输出示例，以及EmoStage各阶段的输出。

\begin{figure}[t]
  \centering
  \includegraphics[width=\linewidth]{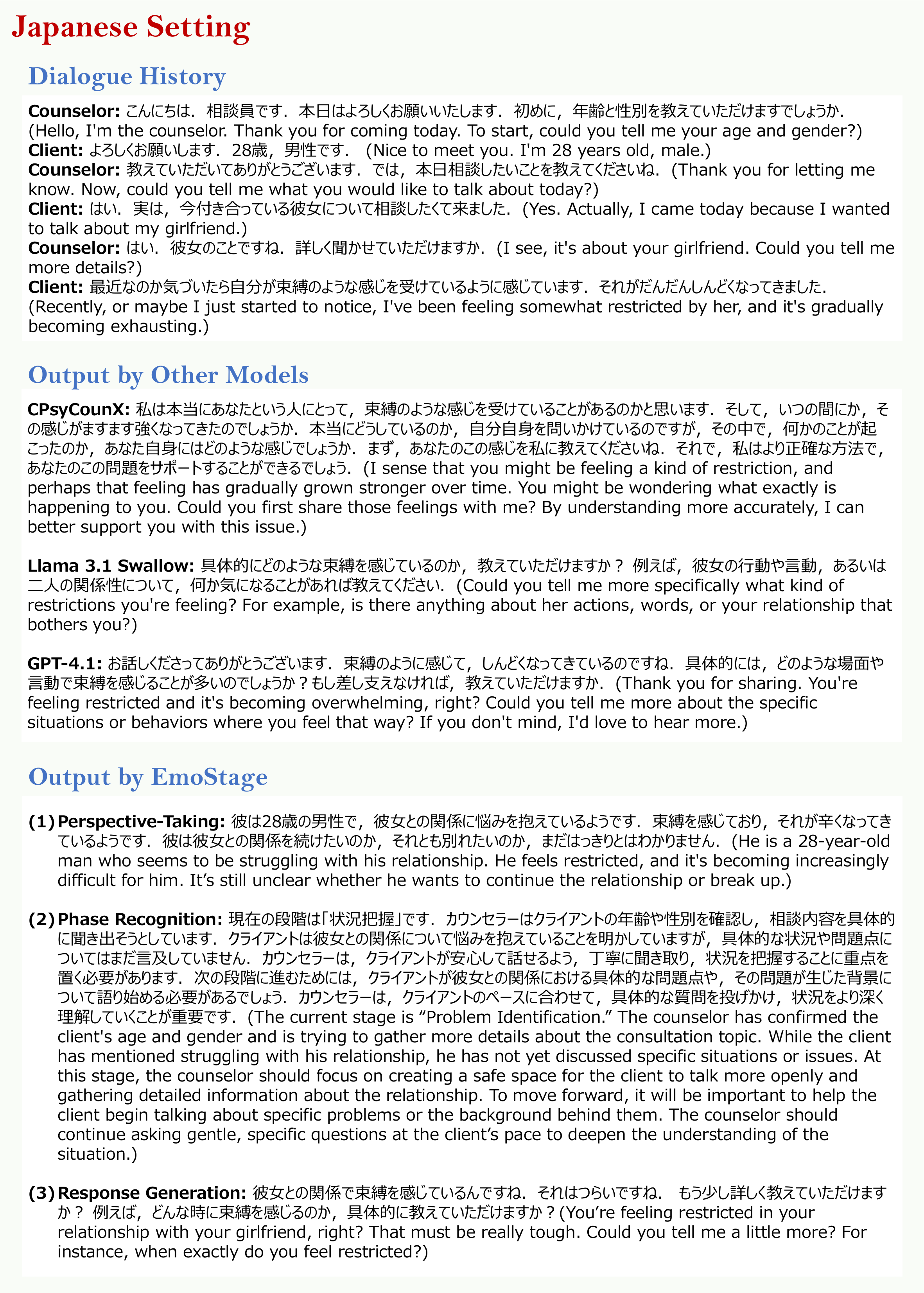}
  \caption{Output examples from each model under the Japanese setting, along with the intermediate outputs from each stage of EmoStage.}
  \label{fig:fig_case_jp}  
\end{figure}

\begin{figure}[t]
  \centering
  \includegraphics[width=\linewidth]{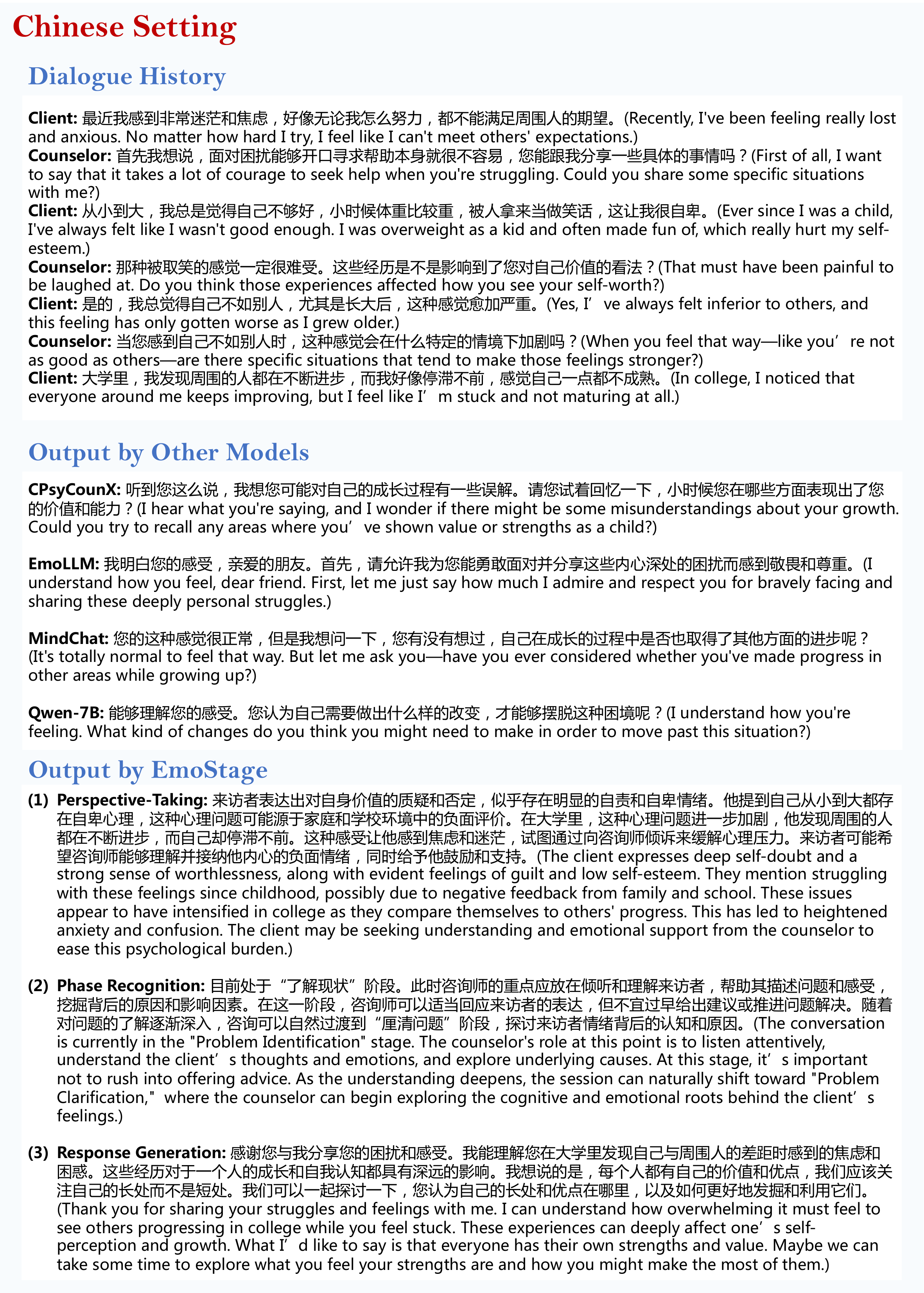}
  \caption{Output examples from each model under the Chinese setting, along with the intermediate outputs from each stage of EmoStage.}
  \label{fig:fig_case_ch}  
\end{figure}

Figure~\ref{fig:fig_case_jp} and Figure~\ref{fig:fig_case_ch} respectively present output examples from each model under the Japanese and Chinese settings, along with the intermediate outputs from each stage of EmoStage.

\end{document}